\pdfoutput=1

\documentclass[11pt]{article}

\usepackage[final]{acl}

\usepackage{times}
\usepackage{latexsym}

\usepackage[T1]{fontenc}

\usepackage[utf8]{inputenc}

\usepackage{microtype}

\usepackage{inconsolata}

\usepackage{maltese}

\usepackage{arabtex}
\usepackage{utf8}
\setcode{utf8}

\usepackage{graphicx}

\usepackage{subcaption}

\usepackage{multirow}

\usepackage{colortbl}

\newcolumntype{C}[1]{>{\centering\let\newline\\\arraybackslash\hspace{0pt}}m{#1}}

\title{Cross-Lingual Transfer from Related Languages:\\
Treating Low-Resource Maltese as Multilingual Code-Switching}

\newcommand*{\authormark}[1][*]{\textsuperscript{#1}}
\author{
    Kurt Micallef\authormark[1]\\
    \texttt{kurt.micallef@um.edu.mt}\\\And
    Nizar Habash\authormark[2]\\
    \texttt{nizar.habash@nyu.edu}\\\And
    Claudia Borg\authormark[1]\\
    \texttt{claudia.borg@um.edu.mt}\\\AND
    Fadhl Eryani\authormark[3,2]\\
    \texttt{fadhl.eryani@nyu.edu}\\\And
    Houda Bouamor\authormark[4]\\
    \texttt{hbouamor@cmu.edu}\\
    \AND 
    {\normalfont\authormark[1]Department of Artificial Intelligence, University of Malta}\\
    {\normalfont\authormark[2]Computational Approaches to Modeling Language Lab, New York University Abu Dhabi}\\
    {\normalfont\authormark[3]University of Tübingen}\\
    {\normalfont\authormark[4]Carnegie Mellon University Qatar}\\
}

\begin{document}
\maketitle
\begin{abstract}
Although multilingual language models exhibit impressive cross-lingual transfer capabilities on unseen languages, the performance on downstream tasks is impacted when there is a script disparity with the languages used in the multilingual model's pre-training data.
Using transliteration offers a straightforward yet effective means to align the script of a resource-rich language with a target language, thereby enhancing cross-lingual transfer capabilities.
However, for mixed languages, this approach is suboptimal, since only a subset of the language benefits from the cross-lingual transfer while the remainder is impeded.
In this work, we focus on Maltese, a Semitic language, with substantial influences from Arabic, Italian, and English, and notably written in Latin script.
We present a novel dataset annotated with word-level etymology.
We use this dataset to train a classifier that enables us to make informed decisions regarding the appropriate processing of each token in the Maltese language.
We contrast indiscriminate transliteration or translation to mixing processing pipelines that only transliterate words of Arabic origin, thereby resulting in text with a mixture of scripts.
We fine-tune the processed data on four downstream tasks and show that conditional transliteration based on word etymology yields the best results, surpassing fine-tuning with raw Maltese or Maltese processed with non-selective pipelines.
\end{list} 
\end{abstract}


\section{Introduction}
\label{section:introduction}

Due to their impressive cross-lingual transfer capabilities, multilingual models have facilitated the development of NLP tools for low-resource languages \cite{kondratyuk-straka-2019-75, wu-dredze-2019-beto, conneau-etal-2020-unsupervised}.
However, multilingual models may fall short in addressing lower-resourced languages \cite{wu-dredze-2020-languages, muller-etal-2021-unseen}. In particular, \citet{muller-etal-2021-unseen} show that the cross-lingual transfer capabilities of a model are affected if the related language seen during pre-training uses a different script.
They further show that transliterating to match the script of the related language improves performance.

In this work, we focus on Maltese -- a Semitic language with an Arabic base and substantial Romance influences written in Latin script.
\citet{micallef-etal-2023-exploring} transliterate Maltese into Arabic script and demonstrate improved performances in certain scenarios when fine-tuning with an Arabic large language model as opposed to a multilingual one in the original script.
However, being influenced by a mixture of languages -- predominantly Arabic, Italian, and English -- we argue that transliterating Maltese entirely into the Arabic script ignores the non-Arabic aspect of the language. Hence, the advantages derived from transliteration are diminished by the losses incurred through moving farther from Italian and English.

Therefore, we propose mixing scripts and applying transliteration selectively.
Specifically, we apply transliteration to Maltese words of Arabic origin, keeping the others in their original Latin script.
We also experiment with mixing transliterations with word-level translations, which yielded the best results overall.

Our main contributions are as follows:
\begin{enumerate}\itemsep0em
    \item
    We annotate a new Maltese dataset with etymological tags (Section~\ref{section:annotations}).
    \item
    We train several etymological classifiers using the annotated data (Section~\ref{section:classifiers}).
    \item
    Using automatic etymological classifications, we define various processing pipelines to conditionally transliterate or translate words based on their etymology (Section~\ref{section:pipelines}).
    \item
    We conduct a thorough evaluation, fine-tuning a variety of language models with different processing pipelines and shed new light on the cross-lingual transfer capabilities exhibited by these models (Section~\ref{section:evaluation}).
\end{enumerate}

The code, the new etymological annotations, and classifiers are released publicly.\footnote{\url{https://github.com/MLRS/malti/tree/2024.eacl}}


\section{Background and Related Work}
\label{section:background}

Due to the mixed nature of Maltese, the language can be viewed as a highly code-switched language among Arabic, Italian, and English.
An analysis of the dictionary from \citet{aquilina-1987-dictionary, aquilina-1990-dictionary} reveals that 32.4\%, 52.5\%, and 6.1\% of Maltese words are of Arabic, Italian/Sicilian, and English origin, respectively \cite{brincat-2017-blending}.
The remaining cases include mixed or unknown-origin words.
We note that Arabic-origin words tend to have higher token frequencies and include function words, and the dictionary entries do not include all inflected cases.

Our work is related to previous works dealing with languages not written in their standard script and/or mixed with other languages, predominantly English.
\citet{pant-dadu-2020-towards} define a pipeline for Hinglish written in Latin script, which only transliterates Hindi-tagged tokens to Devanagari script.
\citet{eskander-etal-2014-foreign} define a pipeline for transliterating Arabizi \cite{darwish-2014-arabizi} text into Arabic script, which includes separate sub-processes for symbols, names, foreign words, and Arabic words.
\citet{shazal-etal-2020-unified} define a neural model for transliteration of Arabizi text into Arabic script, but they skip English words similar to \citet{pant-dadu-2020-towards}.

While these approaches are similar to some of the pipelines presented in Section~\ref{section:pipelines}, the majority of their token distribution (80\%+) is in Latinized Hindi or Arabic, compared to around 60\% Arabic-origin tokens for Maltese (Table~\ref{table:etymology_categories}).
This, in addition to the evolution of Maltese as a distinct language, adds to the complexity of using off-the-shelf models for language modeling \cite{chau-etal-2020-parsing, muller-etal-2021-unseen, micallef-etal-2022-pre}.

Thus, in this work, we build a robust classification model to predict word etymologies, using newly annotated data, to provide more accurate information to our processing pipelines.


\section{Etymology Annotations}
\label{section:annotations}

To build our dataset, we extracted 439 sentences (9,683 tokens) from the Maltese Universal Dependencies Treebank \cite{ceplo-2018-mudt} training set.
We were directly involved in the creation of the guidelines, the annotation of the tokens, including extensive discussions and resolution of disagreements.
Among us, we have native language expertise in Arabic and Maltese and second language expertise in English and Italian.
We relied extensively on authoritative references (mentioned below).
The following are the labels we annotated with.

\paragraph{Arabic}
Maltese tokens of Arabic origin, following the etymological classification by \citet{aquilina-1987-dictionary, aquilina-1990-dictionary}.
This includes words that are derived from Arabic dialects, such as Tunisian Arabic, but we retain the same classification for these.

\paragraph{Non-Arabic}
Maltese tokens whose origin is some language other than Arabic.
During our annotation, we noticed that most of these are of Italian origin.
There were a few cases that were of English origin, for example, \textit{ċekk} `cheque'.
An ambiguity arises for certain Maltese words which correspond to related words in both Italian and English, for example, \textit{rapport} is closely related to both English `report' and Italian `rapporto'.\footnote{
\citet{aquilina-1990-dictionary} lists both Italian and English words as possible cognates for \textit{rapport}.}
Moreover, a few words are also derived from other languages, such as Sicilian and French.
For these reasons, we opt to group these words under this single category.

\paragraph{Mixed}
These are Maltese tokens containing a mixture of Arabic and non-Arabic influences.
The mixed influences take various forms, of which we identify the following sub-categories:

\begin{enumerate}\itemsep0em
    \item\textbf{Verbs}:
    Verbs of non-Arabic origin with Arabic morphology to convey different conjugations.
    For example, \textit{nispjegaw} `we explain', from Italian `spiegare' with the Arabic prefix \textit{ni-} (1\textsuperscript{st}~Person Present) and suffix \textit{-w} (Plural).
    Careful attention was given to Maltese words that share a close surface form with Italian.
    For example, although the Maltese verb \textit{spjega} `he explained' has a similar form to Italian \textit{spiega} `he explains', the difference in their tense inflection lead us to consider the Maltese verb as \textbf{Mixed} and not \textbf{Non-Arabic}, since it does not follow the Italian conjugation rules.
    
    \item\textbf{Plurals}: Non-Arabic-origin nouns that form the plural with Arabic morphology using regular and broken plural formations.
    For example, regular plural \textit{partijiet} `parts' composed of the stem \textit{parti} (Italian `parte'), and the Arabic suffix \textit{-ijiet}; broken plural \textit{ġranet} `days', singular \textit{ġurnata} (Italian `giornata').
    
    \item\textbf{Univerbations}:
    Single words composed of several Arabic and non-Arabic words.
    For example, \textit{minflok} `instead of', which is composed of \textit{minn} `from' (Arabic \<من> \textit{mn}),\footnote{HSB Arabic transliteration \cite{Habash:2007:arabic-transliteration}.} \textit{fi} `in' (Arabic \<في> \textit{fy}), and \textit{lok} `location' (Sicilian `locu', Italian `località').
\end{enumerate}

\paragraph{Code-switching}
Non-Maltese words borrowed from another language, typically English.
As such, these words do not follow Maltese orthographic rules as they are written verbatim from the borrowed words.

\paragraph{Name}
Names of entities that are further categorized into \textbf{Name (Arabic)} and \textbf{Name (Non-Arabic)} for names of Arabic and non-Arabic origin, respectively.
Again, we rely on the etymological classification given by \citet{aquilina-1987-dictionary, aquilina-1990-dictionary}, but make use of additional sources to determine the origin of certain names -- for surnames, for instance, we use Maltagenealogy.\footnote{\url{https://maltagenealogy.com/maltese-surname-origins/}}
Note that this category does not capture entities composed of words that could be used for non-entities.
For example, \textit{Gvern ta' Malta} `Government of Malta' would be considered as a single entity in a Named-Entity Recognition task, but we annotate the phrase as \textit{Gvern}/\textbf{Non-Arabic} \textit{ta'}/\textbf{Arabic} \textit{Malta}/\textbf{Name (Arabic)}.
Non-Maltese words in named entities are tagged as either \textbf{Code-Switching} if translatable, or \textbf{Name} if not. 
For instance, while both words in \textit{Planning Authority} would be classified as \textbf{Code-Switching}, both words in \textit{JF Motors} are tagged as \textbf{Name (Non-Arabic)}.

\paragraph{Symbol}
Tokens that can be considered language universal such as digits and punctuation symbols.

A summary of the annotation frequencies is given in Table~\ref{table:etymology_categories}.
In addition to the raw token counts, we also provide the etymology distribution for the set of unique tokens (types).

\begin{table}[t]
    \centering
    \setlength{\tabcolsep}{2.5pt}
    \begin{tabular}{|l|rr|rr|}
        \hline
        \textbf{Label} & \multicolumn{2}{c|}{\textbf{Token}} &  \multicolumn{2}{c|}{\textbf{Type}}\\
        \hline\hline
        Arabic & 5,848 & 60\% & 1,122 & 47\% \\
        Non-Arabic & 1,559 & 16\% & 660 & 27\% \\
        Mixed & 271 & 3\% & 186 & 8\% \\
        Code-Switching & 398 & 4\% & 169 & 7\% \\
        Name (Arabic) & 146 & 2\% & 36 & 1\% \\
        Name (Non-Arabic) & 423 & 4\% & 171 & 7\% \\
        Symbol & 1,038 & 11\% & 65 & 3\% \\
        \hline
        \textbf{Total} & \textbf{9,683} & \textbf{100\%} & \textbf{2,409} &\textbf{100\%} \\
        \hline
    \end{tabular}
    \caption{Etymology annotation frequencies of tokens and types.}
    \label{table:etymology_categories}
\end{table}


\section{Methodology}
\label{section:methodology}

Our objective is to process Maltese tokens in such a way as to improve cross-lingual transfer.
We design pipelines that use \textbf{transliteration} and \textbf{translation} as our main tools to process Maltese (Section~\ref{section:pipelines}).

For \textbf{transliteration}, we use the implementation from \citet{micallef-etal-2023-exploring}.
Specifically, we extend the non-deterministic character mappings with Tunisian word model ranking and full closed-class token mappings, by making some modifications to the character maps.
Primarily, we add mappings for digits and other common symbols to Arabic script instead of passing them as is.
We also include additional mappings for some letters that were missing in \citet{micallef-etal-2023-exploring}, such as from \textit{t} to \<ث>~{$\theta$}.

For \textbf{translation}, word-level translations are extracted from Google Translate.
Admittedly, this may give sub-optimal translations due to the lack of sentence context.
However, we do not translate at the sentence level because we make token-level decisions and sometimes require partial translations of a subset of words in a sentence.
In addition, most of the tasks used in the evaluation (Section~\ref{section:evaluation}) are token-level classification tasks.
Hence, we decided against using word aligners with a sentence translation since this would amplify the noise in the processing pipeline.
At the same time, word-level translations allow us to reduce the processing power needed, as they are extracted once on the unique set of tokens in the datasets used in Section~\ref{section:evaluation}, and saved as static token mappings.

The processing pipelines make use of an etymology classifier (Section~\ref{section:classifiers}), which also uses the transliterations and the translations as features.

\subsection{Etymology Classifiers}
\label{section:classifiers}

Using the data from Section~\ref{section:annotations}, we build a classifier.
We experiment with the following models.

\paragraph{Translation}
A set of heuristics based on word-level translations and the edit distances between them and the original token.
When the distance with both the Italian and English translations is 0, it is considered to be a Symbol if it contains digits or punctuation symbols and a Name otherwise.
When the distance with either of the Italian or English translations is 0, it is considered to be Code-Switching.
Otherwise, the token is considered to be Arabic or Non-Arabic, based on the minimum distance between the Arabic, Italian, and English translations.
We calculate the Arabic distance using the transliteration instead of the original token.
As such this is not trained on the data as it uses the features statically.

\paragraph{MLE}
A Maximum Likelihood Estimator that predicts the tag observed for the token in the training data.
When multiple tags are observed for a given word, the most frequently seen tag is predicted. If a token has never been encountered before, the most commonly observed tag is predicted, which, in this context, is Arabic.

\paragraph{CRF}
A Conditional Random Field \cite{lafferty-etal-2001-crf} model which makes predictions using the sentence context.
In addition to the original and lower-cased token and positional markers, for each token, the following features are included:
\begin{itemize}\itemsep0em
    \item\textbf{Orthography}:
    low-level boolean features indicating the presence of uppercase characters, digits, punctuation symbols, and Maltese special characters (\textit{ċ}, \textit{ġ}, \textit{\mh}, and \textit{ż}).
    \item\textbf{N-Grams}:
    Boolean features indicating the presence of a frequent n-gram in the token and the presence of each n-gram in the token.
    A set of 197 frequent n-grams is extracted by taking the unique uncased words from Korpus Malti v4.0 \cite{micallef-etal-2022-pre} and computing the most common trigrams and bigrams.
    \item\textbf{Closed-Class}:
    a boolean feature indicating whether the token is one of the full closed-class tokens from \citet{micallef-etal-2023-exploring}.
    \item\textbf{Trans\textsuperscript{2}}: 
    the \textbf{translations} of a token into Arabic, Italian, and English, taken from the pre-computed token-level translations.
    We also include the \textbf{transliteration} of the token into Arabic.
    \item\textbf{Distances}:
    the Levenshtein distance \cite{levenshtein-1966-levenshtein} between the token and each of the translations.
    The Arabic translation distance is computed using the transliteration.
\end{itemize}

All features except for the trans\textsuperscript{2} features, are based on the implementation from \citet{osmelak-wintner-2023-denglisch}.

\paragraph{Ensemble}
We combine MLE and CRF into one model.
This favors the predictions from the MLE model whenever the token is seen exclusively with a single tag.
Otherwise, the predictions from the CRF model are used.

All models are trained using 10-fold cross-validation, using the same splits. The results are shown in Table~\ref{table:etymology_results_full}, reporting the accuracy from all folds.
For the CRF model, we contrast the performance of using no features, adding the orthography features only, adding every other group of features on top of this, and adding all of the features together.
For the Ensemble, we show the results with all features.
In addition to the scores for the entire data, we also show individual results for tokens that are seen in the corresponding training split versus tokens that are not seen.

\begin{table}[t]
    \centering
    
    \begin{subtable}{\linewidth}
        \centering
        \begin{tabular}{|l||c|c|c|}
            \hline
            \textbf{Model} & \textbf{All} & \textbf{Seen} & \textbf{Unseen} \\
            \hline\hline
            Translation & 69.72 & 70.27 & 66.39 \\
            \hline
            MLE & 92.11 & 99.76 & 43.64 \\
            \hline
            CRF & 91.97 & 99.20 & 45.93 \\
            + orthography & 92.90 & 99.22 & 52.64 \\
            ~~~+ n-grams & 96.51 & 99.43 & 78.19 \\
            ~~~+ closed-class & 92.98 & 99.17 & 53.41 \\
            ~~~+ trans$^2$ & 93.77 & 99.36 & 58.49 \\
            ~~~+ distances & 95.75 & 99.29 & 73.35 \\
            ~~~+ all features & 97.55 & 99.64 & \textbf{84.35} \\
            \hline
            Ensemble & \textbf{97.69}& \textbf{99.80} & \textbf{84.35} \\
            \hline
        \end{tabular}
        \vspace{-0.1cm}
        \caption{All Categories ($n=7$)}
        \label{table:etymology_results_full}
        \vspace{0.25cm}
    \end{subtable}

    \begin{subtable}{\linewidth}
        \centering
        \begin{tabular}{|l||c|c|c|}
            \hline
            \textbf{Model} & \textbf{All} & \textbf{Seen} & \textbf{Unseen} \\
            \hline\hline
            Translation & 73.89 & 73.68 & 75.15 \\
            \hline
            MLE & 92.13 & 99.78 & 43.64 \\
            \hline
            CRF & 98.26 & 99.61 & \textbf{89.80} \\
            \hline
            Ensemble & \textbf{98.43} & \textbf{99.81} & \textbf{89.80} \\
            \hline
        \end{tabular}
        \vspace{-0.1cm}
        \caption{Merged Categories ($n=5$)}
        \label{table:etymology_results_merged}
    \end{subtable}
    
    \caption{Etymology classification accuracy across 10-fold cross-validation.}
\end{table}

\begin{figure*}[t]
    \centering
    
    \begin{subfigure}[T]{0.5\linewidth}
        \centering
        \includegraphics[width=\linewidth]{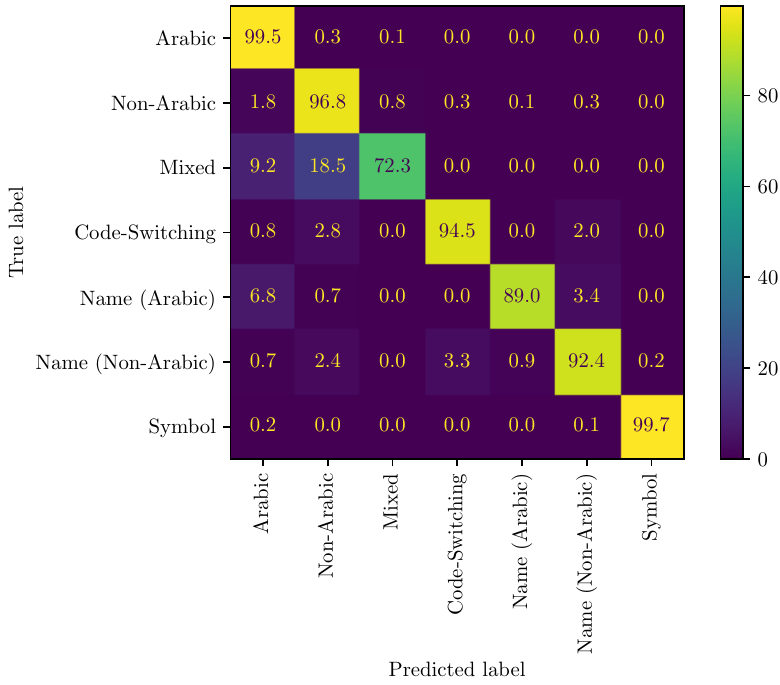}
        \caption{All Categories}
        \label{figure:etymology_confusion_matrix}
    \end{subfigure}
    \hfill
    \begin{subfigure}[T]{0.465\linewidth}
        \centering
        \includegraphics[width=\linewidth]{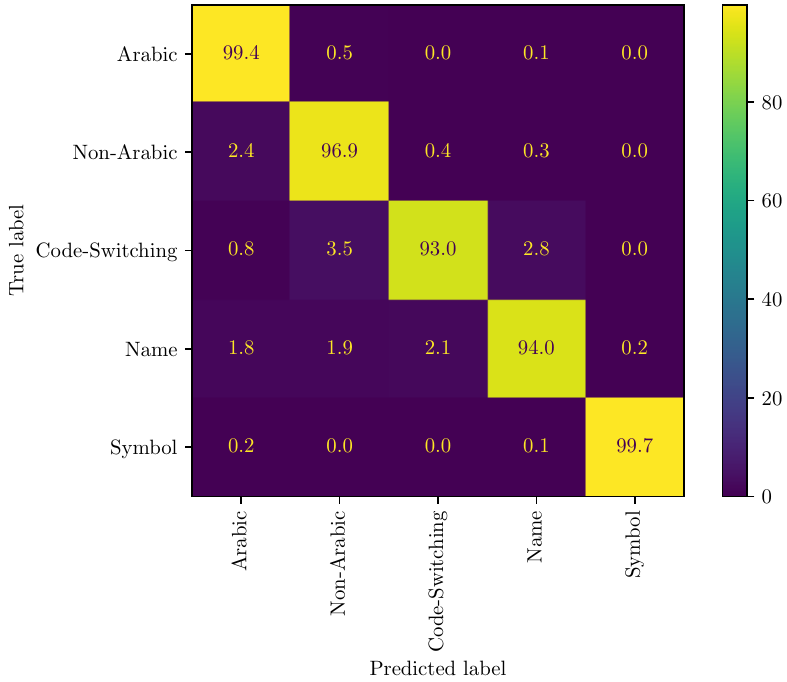}
        \hfill
        \caption{Merged Categories}
        \label{figure:etymology_confusion_matrix_merged}
    \end{subfigure}
    
    \caption{Confusion matrices for the Ensemble classifier.
    Values are percentages and are obtained by adding predictions across folds and normalizing them by dividing by the total for a given class.}
\end{figure*}

With the exception of the Translation model, all models achieve over 91\% accuracy, with Ensemble achieving the best results overall.
While the Translation model performs relatively worse, it performs evenly across seen and unseen tokens.
In contrast, the MLE model is heavily biased towards seen words.
With no additional features, the CRF model performs worse overall than the MLE model, albeit obtaining a higher accuracy on unseen words.
This highlights that many tokens in the data are unambiguous as a simple MLE baseline achieves close to perfect performance on seen tokens.

Adding low-level orthographic features is sufficient for the CRF model to perform better than the MLE baseline.
The other features all contribute to some degree of further improvement, particularly on unseen words which boosts the overall accuracy.
The n-gram and distance features give the most noticeable improvements overall with a 20-25\% improvement on unseen tokens.

All features together yield the best performance for the CRF model.
Despite this, accuracy scores on seen words are worse than those of the MLE model.
This is mitigated by the Ensemble model, which gets an even better score on seen words than MLE, since it gets better predictions on tokens that are seen but with different labels.

Analyzing the predictions reveals that the model makes systematic errors, as shown in Figure~\ref{figure:etymology_confusion_matrix}.
In particular, the Mixed class is considerably mispredicted as non-Arabic or Arabic.
To a lesser degree, Names are conflated with each other, with some confusion with non-name labels, namely, Non-Arabic and Code-Switching for Name (Non-Arabic), and Arabic for Name (Arabic).

\paragraph{Merged Categories} To mitigate the possible negative effect of such mispredictions on our downstream task, we merge the Mixed tag with the \textbf{Non-Arabic} tag, and the Name tags together under a single \textbf{Name} tag. 
The total number of categories is thus reduced from 7 (All) to 5 (Merged).
The decision to merge is motivated by system design and does not invalidate the importance of the various annotated categories, which were driven by linguistic insights.
Furthermore, we note that tokens from the merged categories share a common set of properties 
and merging does not impact the decisions made in Section~\ref{section:pipelines}. 

We report the results of the models using all features in Table~\ref{table:etymology_results_merged}, showing similar trends to Table~\ref{table:etymology_results_full}.
All models attain higher accuracy scores on these merged categories, with the CRF and Ensemble models getting close to 90\% accuracy on unseen words.
As shown in Figure~\ref{figure:etymology_confusion_matrix_merged}, the confusion is drastically reduced overall as well.

Going forward, we use the Ensemble model trained on all the data with merged categories.

\subsection{Text Processing Pipelines}
\label{section:pipelines}

\newcommand{\pass}{\texttt{P}}
\newcommand{\transliterate}{\texttt{X\textsubscript{\ara}}}
\newcommand{\translate}[2]{\texttt{T\rlap{\textsuperscript{#1}}\textsubscript{#2}}}
\newcommand{\mix}[1]{\transliterate/\translate{}{#1}}
\newcommand{\semitransliterate}{\transliterate/\pass}

\newcommand{\mlt}{\texttt{mlt}}
\newcommand{\ara}{\texttt{ara}}
\newcommand{\ita}{\texttt{ita}}
\newcommand{\eng}{\texttt{eng}}
\newcommand{\mul}{\texttt{mul}}

We now make use of the classifier outlined in Section~\ref{section:classifiers} to make decisions on how to process Maltese text.
We define processing pipelines that, given a sequence of Maltese tokens, output another sequence of the same length.
A given token is processed in one of the following ways:
\begin{itemize}\itemsep0em
    \item \textbf{Pass (\pass)}:
    The original token is returned without any modification, so using Maltese as is.
    \item \textbf{Transliteration (\transliterate)}:
    Transliteration into Arabic script.
    \item \textbf{Translation (\translate{src}{tgt})}:
    Translation from a source language \texttt{src} to a target language \texttt{tgt}.
    We consider Arabic (\ara), Italian (\ita), and English (\eng) as different target languages.
\end{itemize}

\newcommand{\multilinecell}[2][c]{%
  \begin{tabular}[#1]{@{}c@{}}#2\end{tabular}}

\begin{table*}[t]
    \footnotesize
    \centering
    \begin{tabular}{|c|ccccccccc|}
        \hline
        
        \multilinecell{\textbf{etymology}\\\textbf{label}} & \multilinecell{Arabic} & \multilinecell{Non-\\Arabic} & \multilinecell{Name} & \multilinecell{Arabic} & \multilinecell{Symbol} & \multilinecell{Arabic} & \multilinecell{Code-\\Switching } & \multilinecell{Non-\\Arabic} & \multilinecell{Symbol} \\
        \hline
        
        \textbf{token} & Il- & karozza & Porsche & tal- & 2022 & g\mh andha & speed & fenomenali & ! \\
        \hline\hline
        \textbf{\pass} & Il- & karozza & Porsche & tal- & 2022 & g\mh andha & speed & fenomenali & ! \\
        
        \hline
        
        \textbf{\transliterate} & \<ال> & \<كردزة> & \<برسكهي> & \<تاع ال> & \<٢٢٠٢> & \<عندها> & \<صباد> & \<فنمنلي> & ! \\
        
        \hline
        \textbf{\translate{}{\ara}} & \<ال> & \<ترام> & \<بورشه> & \<ل> & 2022 & \<هو> & \<سرعة> & \<هائل> & ! \\
        \textbf{\translate{}{\ita}} & IL & tram & Porsche & Di & 2022 & Esso & velocità & fenomenale & ! \\
        \textbf{\translate{}{\eng}} & The & streetcar & Porsche & of & 2022 & it & speed & phenomenal & ! \\
        
        \hline
        
        \textbf{\semitransliterate} & \<ال> & karozza & Porsche & \<تاع ال> & 2022 & \<عندها> & speed & fenomenali & ! \\
        
        \hline
        
        \textbf{\mix{\ara}} & \<ال> & \<ترام> & \<بورشه> & \<تاع ال> & \<٢٢٠٢> & \<عندها> & \<سرعة> & \<هائل> & ! \\
        \textbf{\mix{\ita}} & \<ال> & tram & Porsche & \<تاع ال> & 2022 & \<عندها> & speed & fenomenale & ! \\
        \textbf{\mix{\eng}} & \<ال> & streetcar & Porsche & \<تاع ال> & 2022 & \<عندها> & speed & phenomenal & ! \\
        
        \hline
    \end{tabular}
    \caption{
    An example sentence and the corresponding etymology labels passed through the pipelines outlined in Table~\ref{table:pipelines}.
    The raw sentence is: \textit{Il-karozza Porsche tal-2022 g\mh andha speed fenomenali!} `The 2022 Porsche car has phenomenal speed!'.
    }
    \label{table:example}
\end{table*}

We design several processing pipelines in which we apply one of these actions for a given token, depending on the token's etymology label.
The different pipelines and corresponding actions are summarized in Table~\ref{table:pipelines} and we also show their outputs after processing a sample sentence in Table~\ref{table:example}.

\newcommand{\rotatedcell}[2][l]{%
    \rotatebox[origin=#1]{90}{#2 }}

\begin{table}[t]
    \centering
    \begin{tabular}{|l||l|l|l|l|l|}
        \hline
        
        & \rotatedcell{Arabic} & \rotatedcell{Non-Arabic} & \rotatedcell{Code-Switching} & \rotatedcell{Name} & \rotatedcell{Symbol} \\
        \hline\hline
        
        \textbf{\pass} & \pass & \pass & \pass & \pass & \pass \\
        
        \hline
        
        \textbf{\transliterate} & \transliterate & \transliterate & \transliterate & \transliterate & \transliterate \\
        
        \hline
        
        \textbf{\translate{}{\ara}} & \translate{\mlt}{\ara} & \translate{\mlt}{\ara} & \translate{\mlt}{\ara} & \translate{\mlt}{\ara} & \translate{\mlt}{\ara} \\
        \textbf{\translate{}{\ita}} & \translate{\mlt}{\ita} & \translate{\mlt}{\ita} & \translate{\mlt}{\ita} & \translate{\mlt}{\ita} & \translate{\mlt}{\ita} \\
        \textbf{\translate{}{\eng}} & \translate{\mlt}{\eng} & \translate{\mlt}{\eng} & \translate{\mlt}{\eng} & \translate{\mlt}{\eng} & \translate{\mlt}{\eng} \\
        
        \hline
        
        \textbf{\semitransliterate} & \transliterate & \pass & \pass & \pass & \pass \\
        
        \hline
        
        \textbf{\mix{\ara}} & \transliterate & \translate{\mlt}{\ara} & \translate{\eng}{\ara} & \translate{\mlt}{\ara} & \transliterate \\
        \textbf{\mix{\ita}} & \transliterate & \translate{\mlt}{\ita} & \pass & \translate{\mlt}{\ita} & \pass \\
        \textbf{\mix{\eng}} & \transliterate & \translate{\mlt}{\eng} & \pass & \translate{\mlt}{\eng} & \pass \\
        
        \hline
    \end{tabular}
    \caption{
    Data processing pipelines and the action performed for each corresponding etymology class: transliteration (\transliterate), translation (\translate{src}{tgt}), and pass/nothing (\pass).
    }
    \label{table:pipelines}
\end{table}

The \pass, \transliterate, and \translate{}{*} pipelines perform a pass, transliterate, and translate action indiscriminately, and hence, do not use etymology classifications.

The \mix{*} pipelines mix transliteration and translation.
\mix{\ara} transfers every token to Arabic script by transliterating tokens of Arabic origin and Symbols, translating everything else.
As highlighted in the human evaluation by \citet{micallef-etal-2023-exploring}, transliterations of Maltese words of Arabic origin are generally mapped to the Arabic cognate, whereas the transliteration system does not produce a coherent output for Maltese words of non-Arabic origin.
Thus, we map tokens we expect to be distant from Arabic using translation instead of transliteration.
Differently from the \translate{}{*} pipelines, Code-Switching tokens are translated from English instead of Maltese.\footnote{
We do not consider translating from Italian since almost all cases of code-switching observed during our annotation in Section~\ref{section:annotations} are in English.
}

The \mix{\ita} and \mix{\eng} pipelines similarly mix transliteration with Italian and English translations, respectively.
This produces text that combines a mixture of scripts seamlessly.
Differently from \transliterate/\translate{}{\ara}, we do not translate Code-Switching tokens, since these can already be considered as non-Maltese tokens and the output produced by the \mix{\ita} and \mix{\eng} pipelines already contains a mixture of scripts.
Similarly, the \semitransliterate{} pipeline produces mixed script text by combining transliteration with pass.
The rationale for this pipeline is similar to the \mix{\ita} and \mix{\eng} pipelines.
However, with this pipeline, the aim is to measure the impact of minimizing script differences between related words without using translation.


\section{Downstream Task Evaluation}
\label{section:evaluation}

In this section, we conduct an extrinsic evaluation on four downstream tasks: Part-of-Speech tagging (POS), Dependency Parsing (DP), Named-Entity Recognition (NER), and Sentiment Analysis (SA).
Refer to Section~\ref{section:tasks} for further details on the tasks.

Each dataset is processed using all of the pipelines presented in Section~\ref{section:pipelines}, keeping the corresponding labels/tags the same.
The processed datasets are then used to fine-tune pre-trained language models.
We run fine-tuning 5 times with different random seeds and report the mean performance.
The language models used are the multilingual model mBERT \cite{devlin-etal-2019-bert}, the Arabic model CAMeLBERT-Mix \cite{inoue-etal-2021-interplay}, the Italian model ItalianBERT \cite{italian_bert}, the English model BERT \cite{devlin-etal-2019-bert}, and the Maltese model BERTu \cite{micallef-etal-2022-pre}.

Due to the large number of combinations, we do not fine-tune every model on all the pipelines.
Instead, we only fine-tune models on the pipelines which produce data in a language that it has been intentionally pre-trained on.
So we fine-tune CAMeLBERT on all pipelines which do a \translate{}{\ara} and/or \transliterate{} action, ItalianBERT on all pipelines which do a \translate{}{\ita} action, and BERT on all pipelines which do a \translate{}{\eng} action.
mBERT is fine-tuned on all pipelines since it is multilingual.
Additionally all models are fine-tuned on the \pass{} and \semitransliterate{} pipelines to test their capabilities on Maltese using only fine-tuning data.
The results are presented in Section~\ref{section:results}

\subsection{Tasks}
\label{section:tasks}

We follow all fine-tuning architectures and hyper-parameters suggested by \citet{micallef-etal-2022-pre}.
See Appendix~\ref{appendix:experiments} for further details.

We use the MUDT \cite{ceplo-2018-mudt} dataset for the DP task.
For the POS task, we use the MLRS POS dataset \cite{gatt-ceplo-2013-mlrs} with the same splits from \citet{micallef-etal-2023-exploring}.
The dataset from \citet{martinez-garcia-etal-2021-evaluating} is used for SA, tokenized as in \citet{micallef-etal-2023-exploring} to allow for the token-level actions used to process the data.

We use the MAPA NER data \cite{gianola-2020-mapa} for the NER task using only the level 1 tags.
However, we normalize this data to be in line with the tokenization scheme used in the MUDT and MLRS POS datasets (see Appendix~\ref{appendix:mapa} for further details). 
This step is crucial since the original dataset splits off the \textbf{\textit{-}} and \textbf{\textit{'}} characters as separate tokens.
These characters carry important linguistic features in Maltese which are used by the transliteration system \cite{micallef-etal-2023-exploring} and can at times change the meaning of the token.\footnote{
For example, \textit{fil-} `in the' and \textit{fil} `mortar joint' or \textit{ta'} `of' and \textit{ta} `he gave'.
}

\subsection{Results}
\label{section:results}

The results are summarized in Table~\ref{table:results}.
As expected, BERTu remains the best-performing model across all tasks due to its pre-training on Maltese.
With \pass{}, mBERT performs worse than BERTu. However, it does better than the other monolingual models.
This is largely due to its multilinguality, as it was exposed to several different languages, including those related to Maltese. In contrast, the other monolingual models only include some of the languages with relevance to Maltese.
Moreover, CAMeLBERT performs the worst on the \pass{} pipeline due to the script difference.
Hence, we designate \pass{}~BERTu as the topline setup to close the gap with, and \pass{}~mBERT as the baseline setup to beat.
Similar to the findings from \citet{micallef-etal-2023-exploring}, \transliterate{} CAMeLBERT performs better than \transliterate{} mBERT on POS and SA and \pass{} mBERT on DP and SA.

\begin{table}
    \centering
    
    \begin{subtable}{\linewidth}
        \centering
        \begin{tabular}{|p{1.82cm}||C{0.82cm}|C{0.82cm}|C{0.82cm}|C{0.82cm}|}
            \hline
            \textbf{Pipeline} & \textbf{\textsc{POS}} & \textbf{DP} & \textbf{NER} & \textbf{SA} \\
            & Acc. & LAS & F1 & F1 \\
            \hline\hline
            \pass{} (Topline) & 98.3 & 88.1 & 84.0 & 83.1 \\
            \hline
        \end{tabular}
        \vspace{-0.1cm}
        \caption{BERTu \cite{micallef-etal-2022-pre}}
        \label{table:results_bertu}
        \vspace{0.25cm}
    \end{subtable}
    
    \begin{subtable}{\linewidth}
        \centering
        \begin{tabular}{|p{1.82cm}||C{0.82cm}|C{0.82cm}|C{0.82cm}|C{0.82cm}|}
            \hline
            \textbf{Pipeline} & \textbf{\textsc{POS}} & \textbf{DP} & \textbf{NER} & \textbf{SA} \\
            & Acc. & LAS & F1 & F1 \\
            \hline\hline
            \pass & 88.9 & 61.8 & 75.5 & 61.9 \\
            \hline
            \transliterate & 96.0 & \cellcolor{gray!20}77.3 & 74.1 & \cellcolor{gray!20}67.5 \\
            \hline
            \translate{}{\ara} & 91.2 & 69.6 & 77.2 & \cellcolor{gray!20}73.8 \\
            \hline
            \semitransliterate & 95.6 & 76.6 & 76.9 & 65.0 \\
            \hline
            \mix{\ara} & 95.7 & \cellcolor{gray!20}77.3 & 76.2 & \cellcolor{gray!20}70.2 \\
            \mix{\ita} & 95.5 & 76.5 & 76.2 & 66.2 \\
            \mix{\eng} & 96.0 & \cellcolor{gray!20}77.4 & 78.2 & 64.1 \\
            \hline
        \end{tabular}
        \vspace{-0.1cm}
        \caption{CAMeLBERT-Mix \cite{inoue-etal-2021-interplay}}
        \label{table:results_camelbert}
         \vspace{0.25cm}
    \end{subtable}
    
    \begin{subtable}{\linewidth}
        \centering
        \begin{tabular}{|p{1.82cm}||C{0.82cm}|C{0.82cm}|C{0.82cm}|C{0.82cm}|}
            \hline
            \textbf{Pipeline} & \textbf{\textsc{POS}} & \textbf{DP} & \textbf{NER} & \textbf{SA} \\
            & Acc. & LAS & F1 & F1 \\
            \hline\hline
            \pass & 89.7 & 64.3 & 80.1 & 62.3 \\
            \hline
            \translate{}{\ita} & 92.7 & 71.9 & 79.6 & \cellcolor{gray!20}70.9 \\
            \hline
            \semitransliterate & 44.9 & 14.6 & 76.1 & 58.2 \\
            \hline
            \mix{\ita} & 47.9 & 17.9 & 76.2 & 64.1 \\
            \hline
        \end{tabular}
        \vspace{-0.1cm}
        \caption{ItalianBERT \cite{italian_bert}}
        \label{table:results_italianbert}
         \vspace{0.25cm}
    \end{subtable}
    
    \begin{subtable}{\linewidth}
        \centering
        \begin{tabular}{|p{1.82cm}||C{0.82cm}|C{0.82cm}|C{0.82cm}|C{0.82cm}|}
            \hline
            
            \textbf{Pipeline} & \textbf{\textsc{POS}} & \textbf{DP} & \textbf{NER} & \textbf{SA} \\
            & Acc. & LAS & F1 & F1 \\
            \hline\hline
            \pass & 96.1 & 73.0 & 79.7 & 64.2 \\
            \hline
            \translate{}{\eng} & 93.6 & 74.7 & 79.9 & \cellcolor{gray!20}\textbf{75.2} \\
            \hline
            \semitransliterate & 96.0 & 72.5 & 77.9 & 63.7 \\
            \hline
            \mix{\eng} & 96.4 & 73.9 & 79.1 & \cellcolor{gray!20}69.4 \\
            \hline
        \end{tabular}
        \vspace{-0.1cm}
        \caption{BERT \cite{devlin-etal-2019-bert}}
        \label{table:results_bert}
         \vspace{0.25cm}
    \end{subtable}
    
    \begin{subtable}{\linewidth}
        \centering
        \begin{tabular}{|p{1.82cm}||C{0.82cm}|C{0.82cm}|C{0.82cm}|C{0.82cm}|}
            \hline
            \textbf{Pipeline} & \textbf{\textsc{POS}} & \textbf{DP} & \textbf{NER} & \textbf{SA} \\
            & Acc. & LAS & F1 & F1 \\
            \hline\hline
            \pass{} (Baseline) & 96.7 & 77.3 & 81.0 & 67.3 \\
            \hline
            \transliterate & 95.8 & \cellcolor{gray!20}77.4 & 75.7 & 62.5 \\
            \hline
            \translate{}{\ara} & 91.4 & 71.3 & 77.5 & \cellcolor{gray!20}74.3 \\
            \translate{}{\ita} & 92.6 & 72.9 & 79.6 & \cellcolor{gray!20}71.3 \\
            \translate{}{\eng} & 94.2 & 75.8 & \cellcolor{gray!20}81.5 & \cellcolor{gray!20}73.1 \\
            \hline
            \semitransliterate & 96.6 & \cellcolor{gray!20}78.8 & 80.3 & 66.2 \\
            \hline
            \mix{\ara} & 95.4 & \cellcolor{gray!20}77.4 & 76.5 & 66.2 \\
            \mix{\ita} & 96.5 & \cellcolor{gray!20}\textbf{79.2} & 79.7 & \cellcolor{gray!20}67.3 \\
            \mix{\eng} & \cellcolor{gray!20}\textbf{96.8} & \cellcolor{gray!20}\textbf{79.2} & \cellcolor{gray!20}\textbf{82.2} & \cellcolor{gray!20}67.7 \\
            \hline
        \end{tabular}
        \vspace{-0.1cm}
        \caption{mBERT \cite{devlin-etal-2019-bert}}
        \label{table:results_mbert}
    \end{subtable}
    
    \caption{
    Results using the data processing setups defined in Table~\ref{table:pipelines}, grouped by language model.
    Accuracy, Labelled Attachment Score (LAS), span-based F1, and macro-averaged F1 are reported for the POS, DP, NER, and SA tasks, respectively.
    Each value is an average of~5 runs with different random seeds.
    For each task, the best scores (excluding the Topline) are \textbf{bolded}, and all scores better than the Baseline are \colorbox{gray!20}{shaded}.
    }
    \label{table:results}
\end{table}

A discussion of the other pipelines and their results follows.
Unless explicitly mentioned, we do not include BERTu in the rest of this discussion.

\subsubsection{Translations over Transliterations}
\label{section:translation_vs_transliteration}

Using mBERT, the \translate{}{*} pipelines give worse performance on POS and DP compared to \pass{} and \transliterate{}.
Conversely, the monolingual models generally give better performance on these tasks, with the exception of CAMeLBERT which gives worse performance than \transliterate{}.
However, mBERT performs better overall than the monolingual models with the \translate{}{*} pipelines.
A jump in performance is observed on the NER task, using \translate{}{*} compared to the \transliterate{} pipeline, but only mBERT \translate{}{\eng} gives better performance than mBERT \pass{}.

On the other hand, on the SA task \translate{}{*} give better results than \pass{} and \transliterate{}, regardless of the model used.
In fact, the best scores overall are attained with the \translate{}{*} pipelines for the SA task, with BERT \translate{}{\eng} performing the best across all pipelines.
\translate{}{\eng} is, in general, the best-performing pipeline across all \translate{}{*} pipelines, likely due to the prevalence of Maltese-English parallel data compared to other language pairs,\footnote{
For Maltese, OPUS \cite{tiedemann-nygaard-2004-opus} reports 27.9K, 6.0M, and 34.1M parallel sentences with Arabic, Italian, and English, respectively, at the time of writing.
} which, in turn, results in better translation performance between this language pair compared to other pairs.

These findings highlight that while training with translated data can be an effective solution for low-resource languages, it is largely dependent on the type of task and the performance of the translation model.

This trend is also observed for \mix{\ara}.
For the POS and DP tasks, the added translations give worse performance than \transliterate{} but better performance than \translate{}{\ara} due to the decreased translations.
Conversely, the opposite is true for the NER and SA tasks where \mix{\ara} performs better than \transliterate{} but worse than \translate{}{\ara}.

\subsubsection{Multilingual Models, Multilingual Text}
\label{section:multimodel_multitext}

With mBERT, all \mix{*} pipelines give better performances than the corresponding \translate{}{*} pipelines and \transliterate{} on POS, DP, and NER with the exception of \translate{}{\ara} which performs better than \mix{\ara} on NER.
\mix{\ara} yields the worst results of all \mix{*} pipelines, since, similar to \translate{}{\ara}, this is not fully exploiting the multilinguality aspect of the model.
mBERT \mix{\ita} achieves the best overall performance on the DP task.

\mix{\eng} mBERT achieves better results than \pass{} mBERT on all tasks and achieves the best scores across all pipelines in the POS, DP, and NER tasks.
Besides English being the dominant language in mBERT's pre-training data, we hypothesise that the performance of Maltese-English translation models (as highlighted in Section~\ref{section:translation_vs_transliteration}) also plays a role in this result.
Furthermore, as the gap in performance with BERTu is further reduced, this offers a viable option to further give performance improvements over standard fine-tuning for low-resource languages with similar mixing to Maltese.

Although its pre-training data does not include Maltese, mBERT obtains better results on POS, DP, and NER when trained  with \semitransliterate{} instead of \transliterate{} and \translate{}{*} (except for \translate{}{\eng} on NER).
mBERT \semitransliterate{} also achieves a better score than mBERT \pass{} on DP.
This finding supports the evidence from \citet{muller-etal-2021-unseen} who show that transliteration to the same script as the related language in the pre-training data improves cross-lingual transfer.
Additionally, mBERT \semitransliterate{} performs competitively with mBERT \mix{*}, performing slightly better than \mix{\ara} and \mix{\ita} on the POS and NER tasks.

\subsubsection{Monolingual Models, Multilingual Text}
\label{section:monomodel_multitext}

The trends from Section~\ref{section:translation_vs_transliteration} do not hold entirely for \mix{\ita} and \mix{\eng}.
ItalianBERT with \mix{\ita} performs worse, sometimes significantly, compared to \pass{} and \translate{}{\ita}.
\mix{\eng} BERT performs worse than \translate{}{\eng} on all tasks except POS.

CAMeLBERT generally performs worse with \mix{\ita} than \mix{\ara}, \translate{}{\ara}, and \transliterate{}.
The exceptions are \transliterate{} on NER due to the reduction in performance that we observe when transliterating names into Arabic script (Section~\ref{section:arabic_names}), and \translate{}{\ara} on POS and DP since increasing transliterations and decreasing translations show improved performance for these tasks (Section~\ref{section:translation_vs_transliteration}).

Overall these results make sense since we are giving the respective models less of the type of language they were pre-trained on: transliterations in the case of ItalianBERT and BERT and Italian translations in the case of CAMeLBERT.
Similarly, \semitransliterate{} gives worse performance for the non-Arabic monolingual models, since they were not pre-trained on Maltese, although the discrepancy with \mix{\ara} is negligible.

\subsubsection{Arabic Script on Names}
\label{section:arabic_names}

Overall we observe that changing Name tokens to Arabic script (\transliterate{}, \translate{}{\ara}, and \mix{\ara}) gives among the worst results in the NER tasks.
A big factor for this is the lack of casing information not present in the Arabic script, supported by the findings from \citet{mayhew-etal-2019-ner}.

\subsubsection{Multilingual Presence in Monolingual Models}
\label{section:multilingual_monomodel}

Despite having less Arabic text overall, \mix{\eng} yields the best performance for CAMeLBERT on the POS, DP, and NER tasks.
This could be explained by the presence of the Latin script in CAMeLBERT's pre-training data, which is being exploited by the modeling.

Similarly, BERT has likely seen some Arabic text in its pre-training, since with \mix{\eng} it achieves better performance compared to \translate{}{\eng} on POS.
It is also not significantly worse on the other tasks, particularly when compared to the results of \mix{\ita} ItalianBERT.

These results support the findings by \citet{blevins-zettlemoyer-2022-language} and \citet{muennighoff-etal-2023-crosslingual} who identify that large-scale pre-training corpora contain language contamination, resulting in languages that are unintentionally seen at pre-training.


\section{Conclusion}
\label{section:conclusion}

In this work, we analyze how partially transliterating Maltese has an impact on downstream task performance.
We present a newly annotated dataset with word etymology labels and build classifiers to predict these labels.
Using these classifiers, we design various pipelines to make decisions on which tokens to transliterate or otherwise.

Our evaluation using mBERT shows that by exclusively transliterating words of Arabic origin, downstream task performance improves.
The best results are achieved by mixing transliterations with translations, where including English translations yields better results than fine-tuning on the original data on all tasks.
These findings corroborate with those from \citet{muller-etal-2021-unseen}, but we show this further by only transliterating words that would aid cross-lingual transfer.

Future work should explore language adaptation techniques \cite{chau-etal-2020-parsing, pfeiffer-etal-2020-mad} using the pipelines presented here, to further improve the cross-lingual transfer capabilities of multilingual models.
It is also interesting to apply this method during inference of few-shot and zero-shot settings.
We also hope that our newly annotated dataset can be used as a resource to further support the understanding of Maltese.
While we have reduced the script difference between Maltese and its related languages, other linguistic properties can also impact the cross-lingual performance \cite{philippy-etal-2023-identifying} and future work should investigate these facets.


\section{Limitations}
\label{section:limitations}

The pipelines using translation are limited by the performance of the models used.
We did not systematically evaluate different translation systems to find out the best-performing system.

Although word-level translations allow us to reduce computing requirements, sentence-level translations are bound to produce more accurate translations.
Aligning translations to get the corresponding translation for a word is particularly challenging, especially with varying levels of morphological richness and limited tools for low-resource languages.
Even if these pipelines are combined as a single model that produces the output sentence, this still needs to be aligned to the original token in the data for token classification tasks such as Part-of-Speech tagging, Dependency Parsing, and Named-Entity Recognition.

In our pipelines, we have not treated Names much differently from other tokens.
While some names can be handled by transliteration, especially those of Arabic origin, it is more challenging for others, particularly since many names we annotated use English orthographic rules.

Our findings are also limited to the task results presented here.
As such, a wider variety of tasks, including higher-level semantic tasks, is desirable to verify the generalizability of such a method.
Moreover, having this evaluation on a wider variety of language models would be ideal to assess how factors such as pre-training data and model architecture influence the results.


\section{Ethics Statement}
\label{section:ethics}

The biases present in the data and language models
we used are inherited.
We acknowledge that some performance errors may be due to introduced ambiguities or errors in the techniques we studied.
That said, we do not foresee any major risks.


\section*{Acknowledgements}

We acknowledge support from the LT-Bridge Project (GA 952194) and DFKI for access to the Virtual Laboratory. We further acknowledge funding by Malta Enterprise. 

\bibliography{anthology,custom}

\begin{thebibliography}{32}
\expandafter\ifx\csname natexlab\endcsname\relax\def\natexlab#1{#1}\fi

\bibitem[{Aquilina(1987)}]{aquilina-1987-dictionary}
Joseph Aquilina. 1987.
\newblock \emph{{M}altese-{E}nglish Dictionary Vol. {I}, {A}-{L}}.
\newblock Midsea Books, Valletta, Malta.

\bibitem[{Aquilina(1990)}]{aquilina-1990-dictionary}
Joseph Aquilina. 1990.
\newblock \emph{{M}altese-{E}nglish Dictionary Vol. {II}, {M}-{Z}}.
\newblock Midsea Books, Valletta, Malta.

\bibitem[{Blevins and Zettlemoyer(2022)}]{blevins-zettlemoyer-2022-language}
Terra Blevins and Luke Zettlemoyer. 2022.
\newblock \href {https://doi.org/10.18653/v1/2022.emnlp-main.233} {Language contamination helps explains the cross-lingual capabilities of {E}nglish pretrained models}.
\newblock In \emph{Proceedings of the 2022 Conference on Empirical Methods in Natural Language Processing}, pages 3563--3574, Abu Dhabi, United Arab Emirates. Association for Computational Linguistics.

\bibitem[{Brincat(2017)}]{brincat-2017-blending}
Joseph~M. Brincat. 2017.
\newblock \href {https://doi.org/doi:10.1515/lexi-2017-0011} {{M}altese: blending {S}emitic, {R}omance and {G}ermanic lexemes}.
\newblock \emph{Lexicographica}, 33(2017):207--224.

\bibitem[{{\v{C}}{\'e}pl{\"o}(2018)}]{ceplo-2018-mudt}
Slavom{\'\i}r {\v{C}}{\'e}pl{\"o}. 2018.
\newblock \href {https://bulbul.sk/phd/Text/Slavomir_Ceplo-autoreferat.pdf} {\emph{Constituent order in {M}altese: A quantitative analysis}}.
\newblock Ph.D. thesis, Charles University, Prague.

\bibitem[{Chau et~al.(2020)Chau, Lin, and Smith}]{chau-etal-2020-parsing}
Ethan~C. Chau, Lucy~H. Lin, and Noah~A. Smith. 2020.
\newblock \href {https://doi.org/10.18653/v1/2020.findings-emnlp.118} {Parsing with multilingual {BERT}, a small corpus, and a small treebank}.
\newblock In \emph{Findings of the Association for Computational Linguistics: EMNLP 2020}, pages 1324--1334, Online. Association for Computational Linguistics.

\bibitem[{Conneau et~al.(2020)Conneau, Khandelwal, Goyal, Chaudhary, Wenzek, Guzm{\'a}n, Grave, Ott, Zettlemoyer, and Stoyanov}]{conneau-etal-2020-unsupervised}
Alexis Conneau, Kartikay Khandelwal, Naman Goyal, Vishrav Chaudhary, Guillaume Wenzek, Francisco Guzm{\'a}n, Edouard Grave, Myle Ott, Luke Zettlemoyer, and Veselin Stoyanov. 2020.
\newblock \href {https://doi.org/10.18653/v1/2020.acl-main.747} {Unsupervised cross-lingual representation learning at scale}.
\newblock In \emph{Proceedings of the 58th Annual Meeting of the Association for Computational Linguistics}, pages 8440--8451, Online. Association for Computational Linguistics.

\bibitem[{Darwish(2014)}]{darwish-2014-arabizi}
Kareem Darwish. 2014.
\newblock \href {https://doi.org/10.3115/v1/W14-3629} {{A}rabizi detection and conversion to {A}rabic}.
\newblock In \emph{Proceedings of the {EMNLP} 2014 Workshop on {A}rabic Natural Language Processing ({ANLP})}, pages 217--224, Doha, Qatar. Association for Computational Linguistics.

\bibitem[{Devlin et~al.(2019)Devlin, Chang, Lee, and Toutanova}]{devlin-etal-2019-bert}
Jacob Devlin, Ming-Wei Chang, Kenton Lee, and Kristina Toutanova. 2019.
\newblock \href {https://doi.org/10.18653/v1/N19-1423} {{BERT}: Pre-training of deep bidirectional transformers for language understanding}.
\newblock In \emph{Proceedings of the 2019 Conference of the North {A}merican Chapter of the Association for Computational Linguistics: Human Language Technologies, Volume 1 (Long and Short Papers)}, pages 4171--4186, Minneapolis, Minnesota. Association for Computational Linguistics.

\bibitem[{Eskander et~al.(2014)Eskander, Al-Badrashiny, Habash, and Rambow}]{eskander-etal-2014-foreign}
Ramy Eskander, Mohamed Al-Badrashiny, Nizar Habash, and Owen Rambow. 2014.
\newblock \href {https://doi.org/10.3115/v1/W14-3901} {Foreign words and the automatic processing of {A}rabic social media text written in {R}oman script}.
\newblock In \emph{Proceedings of the First Workshop on Computational Approaches to Code Switching}, pages 1--12, Doha, Qatar. Association for Computational Linguistics.

\bibitem[{Gatt and {\v{C}}{\'e}pl{\"o}(2013)}]{gatt-ceplo-2013-mlrs}
Albert Gatt and Slavom{\'\i}r {\v{C}}{\'e}pl{\"o}. 2013.
\newblock \href {https://www.bulbul.sk/writings/CL2013-expanded-abstract-AG_SC.pdf} {{Digital Corpora and Other Electronic Resources for Maltese}}.
\newblock In \emph{Proceedings of the International Conference on Corpus Linguistics}, pages 96--97. {UCREL}, Lancaster, UK.

\bibitem[{Gianola et~al.(2020)Gianola, Ēriks Ajausks, Arranz, Bendahman, Bié, Borg, Cerdà, Choukri, Cuadros, de~Gibert, Degroote, Edelman, Etchegoyhen, Ángela Franco~Torres, Hernandez, Pablos, Gatt, Grouin, Herranz, Kohan, Lavergne, Melero, Paroubek, Rigault, Rosner, Rozis, van~der Plas, Vīksna, and Zweigenbaum}]{gianola-2020-mapa}
Lucie Gianola, Ēriks Ajausks, Victoria Arranz, Chomicha Bendahman, Laurent Bié, Claudia Borg, Aleix Cerdà, Khalid Choukri, Montse Cuadros, Ona de~Gibert, Hans Degroote, Elena Edelman, Thierry Etchegoyhen, Ángela Franco~Torres, Mercedes~García Hernandez, Aitor~García Pablos, Albert Gatt, Cyril Grouin, Manuel Herranz, Alejandro~Adolfo Kohan, Thomas Lavergne, Maite Melero, Patrick Paroubek, Mickaël Rigault, Mike Rosner, Roberts Rozis, Lonneke van~der Plas, Rinalds Vīksna, and Pierre Zweigenbaum. 2020.
\newblock \href {https://doi.org/10.3233/FAIA200869} {Automatic removal of identifying information in official eu languages for public administrations: The {MAPA} project}.
\newblock In \emph{Proceedings of the 33rd International Conference on Legal Knowledge and Information Systems ({JURIX'20})}, pages 223--226. IOS Press.

\bibitem[{Habash et~al.(2007)Habash, Soudi, and Buckwalter}]{Habash:2007:arabic-transliteration}
Nizar Habash, Abdelhadi Soudi, and Tim Buckwalter. 2007.
\newblock {On {{A}rabic} Transliteration}.
\newblock In A.~van~den Bosch and A.~Soudi, editors, \emph{{A}rabic Computational Morphology: Knowledge-based and Empirical Methods}, pages 15--22. Springer, Netherlands.

\bibitem[{Inoue et~al.(2021)Inoue, Alhafni, Baimukan, Bouamor, and Habash}]{inoue-etal-2021-interplay}
Go~Inoue, Bashar Alhafni, Nurpeiis Baimukan, Houda Bouamor, and Nizar Habash. 2021.
\newblock \href {https://aclanthology.org/2021.wanlp-1.10} {The interplay of variant, size, and task type in {A}rabic pre-trained language models}.
\newblock In \emph{Proceedings of the Sixth Arabic Natural Language Processing Workshop}, pages 92--104, Kyiv, Ukraine (Virtual). Association for Computational Linguistics.

\bibitem[{Kondratyuk and Straka(2019)}]{kondratyuk-straka-2019-75}
Dan Kondratyuk and Milan Straka. 2019.
\newblock \href {https://doi.org/10.18653/v1/D19-1279} {75 languages, 1 model: Parsing {U}niversal {D}ependencies universally}.
\newblock In \emph{Proceedings of the 2019 Conference on Empirical Methods in Natural Language Processing and the 9th International Joint Conference on Natural Language Processing (EMNLP-IJCNLP)}, pages 2779--2795, Hong Kong, China. Association for Computational Linguistics.

\bibitem[{Lafferty et~al.(2001)Lafferty, McCallum, and Pereira}]{lafferty-etal-2001-crf}
John~D. Lafferty, Andrew McCallum, and Fernando C.~N. Pereira. 2001.
\newblock \href {https://dl.acm.org/doi/10.5555/645530.655813} {Conditional random fields: Probabilistic models for segmenting and labeling sequence data}.
\newblock In \emph{Proceedings of the Eighteenth International Conference on Machine Learning}, ICML '01, pages 282--289, {San Francisco}, {CA}, {USA}. Morgan Kaufmann Publishers Inc.

\bibitem[{Levenshtein(1966)}]{levenshtein-1966-levenshtein}
Vladimir~I. Levenshtein. 1966.
\newblock Binary codes capable of correcting deletions, insertions, and reversals.
\newblock \emph{Soviet Physics -- Doklady}, 10:707--710.

\bibitem[{Mart{\'\i}nez-Garc{\'\i}a et~al.(2021)Mart{\'\i}nez-Garc{\'\i}a, Badia, and Barnes}]{martinez-garcia-etal-2021-evaluating}
Antonio Mart{\'\i}nez-Garc{\'\i}a, Toni Badia, and Jeremy Barnes. 2021.
\newblock \href {https://doi.org/10.18653/v1/2021.acl-long.244} {Evaluating morphological typology in zero-shot cross-lingual transfer}.
\newblock In \emph{Proceedings of the 59th Annual Meeting of the Association for Computational Linguistics and the 11th International Joint Conference on Natural Language Processing (Volume 1: Long Papers)}, pages 3136--3153, Online. Association for Computational Linguistics.

\bibitem[{Mayhew et~al.(2019)Mayhew, Tsygankova, and Roth}]{mayhew-etal-2019-ner}
Stephen Mayhew, Tatiana Tsygankova, and Dan Roth. 2019.
\newblock \href {https://doi.org/10.18653/v1/D19-1650} {{ner and pos when nothing is capitalized}}.
\newblock In \emph{Proceedings of the 2019 Conference on Empirical Methods in Natural Language Processing and the 9th International Joint Conference on Natural Language Processing (EMNLP-IJCNLP)}, pages 6256--6261, Hong Kong, China. Association for Computational Linguistics.

\bibitem[{Micallef et~al.(2023)Micallef, Eryani, Habash, Bouamor, and Borg}]{micallef-etal-2023-exploring}
Kurt Micallef, Fadhl Eryani, Nizar Habash, Houda Bouamor, and Claudia Borg. 2023.
\newblock \href {https://doi.org/10.18653/v1/2023.cawl-1.4} {Exploring the impact of transliteration on {NLP} performance: Treating {M}altese as an {A}rabic dialect}.
\newblock In \emph{Proceedings of the Workshop on Computation and Written Language (CAWL 2023)}, pages 22--32, Toronto, Canada. Association for Computational Linguistics.

\bibitem[{Micallef et~al.(2022)Micallef, Gatt, Tanti, van~der Plas, and Borg}]{micallef-etal-2022-pre}
Kurt Micallef, Albert Gatt, Marc Tanti, Lonneke van~der Plas, and Claudia Borg. 2022.
\newblock \href {https://doi.org/10.18653/v1/2022.deeplo-1.10} {Pre-training data quality and quantity for a low-resource language: New corpus and {BERT} models for {M}altese}.
\newblock In \emph{Proceedings of the Third Workshop on Deep Learning for Low-Resource Natural Language Processing}, pages 90--101, Hybrid. Association for Computational Linguistics.

\bibitem[{Muennighoff et~al.(2023)Muennighoff, Wang, Sutawika, Roberts, Biderman, Le~Scao, Bari, Shen, Yong, Schoelkopf, Tang, Radev, Aji, Almubarak, Albanie, Alyafeai, Webson, Raff, and Raffel}]{muennighoff-etal-2023-crosslingual}
Niklas Muennighoff, Thomas Wang, Lintang Sutawika, Adam Roberts, Stella Biderman, Teven Le~Scao, M~Saiful Bari, Sheng Shen, Zheng~Xin Yong, Hailey Schoelkopf, Xiangru Tang, Dragomir Radev, Alham~Fikri Aji, Khalid Almubarak, Samuel Albanie, Zaid Alyafeai, Albert Webson, Edward Raff, and Colin Raffel. 2023.
\newblock \href {https://doi.org/10.18653/v1/2023.acl-long.891} {Crosslingual generalization through multitask finetuning}.
\newblock In \emph{Proceedings of the 61st Annual Meeting of the Association for Computational Linguistics (Volume 1: Long Papers)}, pages 15991--16111, Toronto, Canada. Association for Computational Linguistics.

\bibitem[{Muller et~al.(2021)Muller, Anastasopoulos, Sagot, and Seddah}]{muller-etal-2021-unseen}
Benjamin Muller, Antonios Anastasopoulos, Beno{\^\i}t Sagot, and Djam{\'e} Seddah. 2021.
\newblock \href {https://doi.org/10.18653/v1/2021.naacl-main.38} {When being unseen from m{BERT} is just the beginning: Handling new languages with multilingual language models}.
\newblock In \emph{Proceedings of the 2021 Conference of the North American Chapter of the Association for Computational Linguistics: Human Language Technologies}, pages 448--462, Online. Association for Computational Linguistics.

\bibitem[{Osmelak and Wintner(2023)}]{osmelak-wintner-2023-denglisch}
Doreen Osmelak and Shuly Wintner. 2023.
\newblock \href {https://doi.org/10.18653/v1/2023.sigtyp-1.5} {The denglisch corpus of {G}erman-{E}nglish code-switching}.
\newblock In \emph{Proceedings of the 5th Workshop on Research in Computational Linguistic Typology and Multilingual NLP}, pages 42--51, Dubrovnik, Croatia. Association for Computational Linguistics.

\bibitem[{Pant and Dadu(2020)}]{pant-dadu-2020-towards}
Kartikey Pant and Tanvi Dadu. 2020.
\newblock \href {https://aclanthology.org/2020.aacl-srw.6} {Towards code-switched classification exploiting constituent language resources}.
\newblock In \emph{Proceedings of the 1st Conference of the Asia-Pacific Chapter of the Association for Computational Linguistics and the 10th International Joint Conference on Natural Language Processing: Student Research Workshop}, pages 37--43, Suzhou, China. Association for Computational Linguistics.

\bibitem[{Pfeiffer et~al.(2020)Pfeiffer, Vuli{\'c}, Gurevych, and Ruder}]{pfeiffer-etal-2020-mad}
Jonas Pfeiffer, Ivan Vuli{\'c}, Iryna Gurevych, and Sebastian Ruder. 2020.
\newblock \href {https://doi.org/10.18653/v1/2020.emnlp-main.617} {{MAD-X}: {A}n {A}dapter-{B}ased {F}ramework for {M}ulti-{T}ask {C}ross-{L}ingual {T}ransfer}.
\newblock In \emph{Proceedings of the 2020 Conference on Empirical Methods in Natural Language Processing (EMNLP)}, pages 7654--7673, Online. Association for Computational Linguistics.

\bibitem[{Philippy et~al.(2023)Philippy, Guo, and Haddadan}]{philippy-etal-2023-identifying}
Fred Philippy, Siwen Guo, and Shohreh Haddadan. 2023.
\newblock \href {https://doi.org/10.18653/v1/2023.sigtyp-1.3} {Identifying the correlation between language distance and cross-lingual transfer in a multilingual representation space}.
\newblock In \emph{Proceedings of the 5th Workshop on Research in Computational Linguistic Typology and Multilingual NLP}, pages 22--29, Dubrovnik, Croatia. Association for Computational Linguistics.

\bibitem[{Schweter(2020)}]{italian_bert}
Stefan Schweter. 2020.
\newblock \href {https://doi.org/10.5281/zenodo.4263142} {{I}talian {BERT} and {ELECTRA} models}.

\bibitem[{Shazal et~al.(2020)Shazal, Usman, and Habash}]{shazal-etal-2020-unified}
Ali Shazal, Aiza Usman, and Nizar Habash. 2020.
\newblock \href {https://aclanthology.org/2020.wanlp-1.15} {A unified model for {A}rabizi detection and transliteration using sequence-to-sequence models}.
\newblock In \emph{Proceedings of the Fifth Arabic Natural Language Processing Workshop}, pages 167--177, Barcelona, Spain (Online). Association for Computational Linguistics.

\bibitem[{Tiedemann and Nygaard(2004)}]{tiedemann-nygaard-2004-opus}
J{\"o}rg Tiedemann and Lars Nygaard. 2004.
\newblock \href {http://www.lrec-conf.org/proceedings/lrec2004/pdf/320.pdf} {The {OPUS} corpus - parallel and free: \url{http://logos.uio.no/opus}}.
\newblock In \emph{Proceedings of the Fourth International Conference on Language Resources and Evaluation ({LREC}{'}04)}, Lisbon, Portugal. European Language Resources Association (ELRA).

\bibitem[{Wu and Dredze(2019)}]{wu-dredze-2019-beto}
Shijie Wu and Mark Dredze. 2019.
\newblock \href {https://doi.org/10.18653/v1/D19-1077} {Beto, bentz, becas: The surprising cross-lingual effectiveness of {BERT}}.
\newblock In \emph{Proceedings of the 2019 Conference on Empirical Methods in Natural Language Processing and the 9th International Joint Conference on Natural Language Processing (EMNLP-IJCNLP)}, pages 833--844, Hong Kong, China. Association for Computational Linguistics.

\bibitem[{Wu and Dredze(2020)}]{wu-dredze-2020-languages}
Shijie Wu and Mark Dredze. 2020.
\newblock \href {https://doi.org/10.18653/v1/2020.repl4nlp-1.16} {Are all languages created equal in multilingual {BERT}?}
\newblock In \emph{Proceedings of the 5th Workshop on Representation Learning for NLP}, pages 120--130, Online. Association for Computational Linguistics.

\end{thebibliography}
\appendix


\section{MAPA Data Fixes}
\label{appendix:mapa}

The Maltese data from \citet{gianola-2020-mapa} is fixed to have consistent tokenization with the other token classification datasets used in Section~\ref{section:evaluation}.
We do this by re-tokenizing the raw text using the MLRS Tokenizer.\footnote{\url{https://mlrs.research.um.edu.mt/}}
Further to this, we also manually split off trailing \textit{-} and \textit{'} for tokens that do not carry the linguistic meaning for Maltese.
For instance, marking number ranges with \textit{-} or using \textit{'} for quotation marks.

Since some of the tokens from the original data are merged into a single token, the corresponding labels are also merged.
Whenever the merged tokens contain different target labels, we keep them separate.

While doing this process, we went through the inconsistencies between the tokens in the data and the new tokens.
While there were legitimate cases where the source tokenization made sense, we identified certain entity spans that were incorrectly marked, typically a missing character in the whole word.
In these cases, we fix the annotation so that the span is consistent with the tokenization.

Lastly, we also fixed some of the labels which contained errors during our conversion.
For cases where the entity span was marked but no label was present, we added the labels.
When there were inconsistencies between the level 1 and level 2 tags, we fixed the incorrect tag appropriately.

We make this dataset publicly available.\footnote{\url{https://huggingface.co/datasets/MLRS/mapa_maltese}}


\section{Experimental Setup}
\label{appendix:experiments}

The number of parameters for language models used in Section~\ref{section:evaluation} is summarized in Table~\ref{table:parameters}.

\begin{table}[ht]
    \centering
    \begin{tabular}{|l|c|}
        \hline
        \textbf{Model} & \textbf{Parameters} \\
        \hline
        BERTu & 126M \\
        mBERT & 179M \\
        CAMeLBERT & 109M \\
        ItalianBERT & 111M \\
        BERT & 109M \\
        \hline
    \end{tabular}
    \caption{Number of parameters for the language models used in Section~\ref{section:evaluation}}
    \label{table:parameters}
\end{table}

We use NVIDIA A100 GPUs (40GB and 80GB, depending on memory requirements) on a compute cluster.
Fine-tuning time depends on the model used and the pipeline from Section~\ref{section:pipelines} with which the data was processed with, but a single GPU was always used.
Giving a rough estimate for each task: Part-of-Speech tagging takes around an hour and a half, Dependency Parsing takes around 1 hour, Named-Entity Recognition takes around 6 hours, and Sentiment Analysis takes around 30 minutes.
Named-Entity Recognition takes significantly longer since the dataset used is larger and we use gradient accumulation to ease memory requirements while keeping the same effective batch size from \citet{micallef-etal-2022-pre}.
The figures reported here include all of the runs with different random seeds, the test evaluation for each run, and any initial setup necessary for startup.

\end{document}